# A Software-equivalent SNN Hardware using RRAM-array for Asynchronous Real-time Learning

A. Shukla[†], V. Kumar, U. Ganguly*
Dept. of Electrical Engineering
Indian Institute of Technology Bombay
Mumbai, India - 400076
[†]adityashukla@ee.iitb.ac.in, *udayan@ee.iitb.ac.in

***Abstract*—Spiking Neural Network (SNN) naturally inspires hardware implementation as it is based on biology. For learning, spike time dependent plasticity (STDP) may be implemented using an energy efficient waveform superposition on memristor based synapse. However, system level implementation has three challenges. First, a classic dilemma is that recognition requires current reading for *short* voltage-spikes which is disturbed by *large* voltage-waveforms that are *simultaneously* applied on the same memristor for real-time learning i.e. the *simultaneous read-write dilemma*. Second, the hardware needs to exactly replicate software implementation for *easy* adaptation of algorithm to hardware. Third, the devices used in hardware simulations must be realistic. In this paper, we present an approach to address the above concerns. First, the learning and recognition occurs in separate arrays simultaneously in real-time, asynchronously – avoiding non-biomimetic clocking based complex signal management. Second, we show that the hardware emulates software at *every* stage by comparison of SPICE (circuit-simulator) with MATLAB® (mathematical SNN algorithm implementation in software) implementations. As an example, the hardware shows 97.5% accuracy in classification which is equivalent to software for a *Fisher's Iris* dataset. Third, the STDP is implemented using a model of synaptic device implemented using $HfO_2$ memristor. We show that an increasingly realistic memristor model slightly reduces the hardware performance (85%), which highlights the need to engineer RRAM characteristics specifically for SNN.***

***Keywords*—*Spiking neural network, synaptic-time-dependent-plasticity, RRAM, memristors, RRAM cross-point array, SPICE, Fischer Iris dataset, Iris classification***

## I. INTRODUCTION

Today, various applications (like voice & image recognition etc.) using brain-like networks are implemented on von-Neumann machines [1]. However, achieving energy efficiency comparable to biology is a major challenge [2]. Hence, the more biomimetic approach Spiking Neural Networks (SNN) promises higher energy efficiency, which motivates hardware implementation. SNN algorithm involves two activities – recognition and learning. In the recognition stage, the $n$-input spikes (input vector) are converted into $m$-output spikes (vector) through the $m \times n$ synaptic-array (see Fig. 1(a)), which is akin to matrix-multiplication. If the synaptic array has the correct synaptic weights, the desired input to output mapping is achieved which leads to recognition. In the learning stage, correct synaptic weights are learnt through a learning rule. Spike-Time-Dependent Plasticity (STDP) is one such learning rule that requires that the synaptic weights are modified ($\Delta w$) based on time correlation ($\Delta t = t_{post} - t_{pre}$) between spike times of pre-synaptic ($t_{pre}$) and post-synaptic ($t_{post}$) neurons (Fig. 1(b)). These pre- and post-synaptic spikes are generated in the recognition stage. *Hence, recognition process must simultaneously occur for learning.* However, for achieving only recognition, once the corrected synaptic weights are set-up in the synaptic array, then no further weight modification is needed and only recognition process is sufficient. In software, both learning and recognition has been demonstrated in [3],[4] and [5]. In hardware, recognition requires static synaptic weights. However learning involves simultaneous recognition (static) and learning (dynamic). Hence, software based learning followed by hardware based recognition is implemented as the first stage – known as *on-chip recognition with off-chip learning.* This has been demonstrated with *binary* SRAM based synaptic array [2]. Further biomimetic *analog* memristor/RRAM devices e.g. 2-PCM synapse [6], Mn doped $HfO_2$ [7], and $Pr_{0.7}Ca_{0.3}MnO_3$ [8]-[9] based RRAMs have replaced binary SRAM. Using such arrays, character recognition was demonstrated and effect of RRAM's stochasticity was studied in [10].

In the second stage, *on-chip recognition and learning* requires that the memristor array be static during recognition and dynamic during learning. Recognition is akin to read where the memristor conductance is static – and requires low bias pulse. Learning is akin to write where memristor conductance changes – and requires large bias. This, *simultaneous read - write dilemma is a challenge* as during read at low bias – conductance cannot change for write; and during write at high bias – conductance is not static for good read. To break this challenge, several digital clock-based scheme where various phases of the clock are used for recognition, communication and weight update akin to time-division-multiplexing access (TDMA) for SRAM [11], [12] and memristor [13] have been developed. In comparison to digital implementation, analog implementation has better area and energy efficiency [14]. A spiking neuron would initiate a voltage-waveform in driver circuits [15], [16].

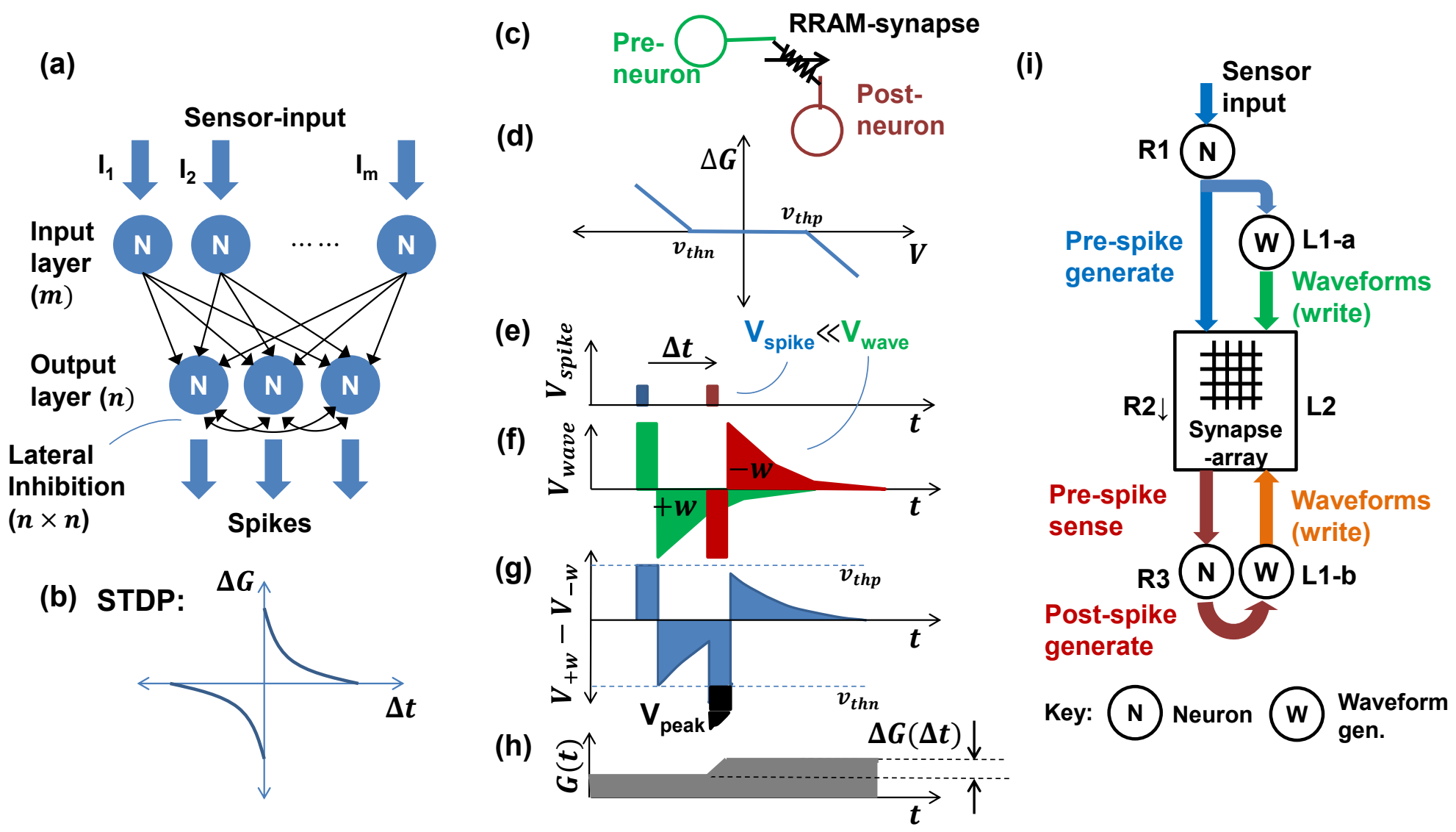

Figure 1: (a) Signal flow diagram for a two layered (m by n) SNN; (b) Synaptic-time dependent plasticity rule (c) RRAM as model of synapse between two neurons; (d) For an ideal RRAM, the conductance change is proportional to the extent by which pulse height exceeds the threshold; below the threshold, no change occurs; (e) Pre- and post-synaptic neuron's pulse (blue and red respectively), (f) respective waveforms, (g) their superposition, and (h) corresponding change in weight of synapse; (i) One-array implementation of a hardware SNN implementation.

A pre-neuron is connected to post-neuron via a synapse (Fig. 1(c)). A RRAM or memristor (used interchangeably) shows voltage-dependent conductance change ($\Delta G$) in both polarity when $V$ exceed a threshold (Fig. 1(d)). As pre- and post- neurons spike (small & short $V_{spike}$; see Fig. 1(e)), corresponding waveforms (high & long $V_{wave}$) are generated (see Fig. 1(f)). The superposition of waveforms generated by pre- and post-synaptic neurons would produce a peak voltage ($V_{peak}$) at the memristor based synapse, which is proportional to the time-correlation ($\Delta t = t_{post} - t_{pre}$) as shown in Fig. 1(g). The $V_{peak}$ produces conductance change ($\Delta G$) at the memristor to enable STDP (Fig. 1(h)). However, the driver based "strong and long" voltage-waveform for STDP based learning will interfere with "weak and short" voltage-spike for recognition to confuse the output neurons. This is a major dilemma for analog, asynchronous, real-time learning in memristor arrays (Fig. 1(i)). In this paper, we present a scheme to implement an asynchronous scheme for a spiking neural network, using two-RRAM arrays to resolve the dilemma. Then, this scheme is demonstrated in a circuit simulation with SPICE for real-time learning. The equivalence of hardware to software is evaluated *microscopically* at every stage pulsing as well as at the compare *macroscopic* performance on a standard *Fisher's Iris* dataset. Finally, the effect of increasingly realistic RRAM circuit model on learning performance is presented.

## II. Software-Equivalent Hardware

The software implementation of a spiking neural network typically consists of (1) neuro-synaptic network (2) neuron model (3) synapse's response to a pre-neuron's spike (that will be proportional to the synapse's strength) and (4) a weight update rule [17]. Using variants of these components, powerful spiking neural networks can be designed.

*a) Network:* There are $m$ input neurons connected to $n$ output neurons through a network on $m \times n$ synaptic array as shown in Fig 1(a).

*b) Neuron Model:* Several models of neurons exist, simplest being the leaky-integrate and fire (LIF) [17]. The LIF neuron state variable varies in time as Eq. 1, where, $V(t)$ is the state variable, $R$ and $C$ are neuron-specific parameters and $I_i(t)$, $i = 1,2..n$ are the excitation sources. Whenever the state variable $V(t)$ reaches a constant threshold ($V_t$), it spikes via an internal mechanism, after which it is reset to a constant value for a constant period (called the refractory period). During this period, the input excitation signal is unable to cause any change in neuron's state.

$$C\frac{dV_j}{dt} = -\frac{V_j(t)}{R} + \sum_i I_{\alpha,i\to j}(t) \qquad (1)$$

*c) Synaptic Response to Spike:* Synapse generates a time varying current in response to its source (pre-) neuron's spike, which may cause/inhibit spiking of the sink (post-) synaptic neuron (See Eq. 2). This current is proportional to the strength/weight ($w_{ij}$) of a synapse beween i$^{th}$ pre-neuron and j$^{th}$ post neuron.

$$I_{\alpha,i\to j}(t) = w_{ij}V_0\left(\exp\left(-\frac{t}{\tau_1}\right) - \exp\left(-\frac{t}{\tau_2}\right)\right) \qquad (2)$$

Here, $V_0, \tau_1$ and $\tau_2$ are network design parameters.

*d) Synaptic weight update rule:* Depending on the timings of the pre- and post-synaptic neurons, the synaptic weight changes according to a network specific rule. For all purposes, STDP rule given in Eq. 3 has been used.

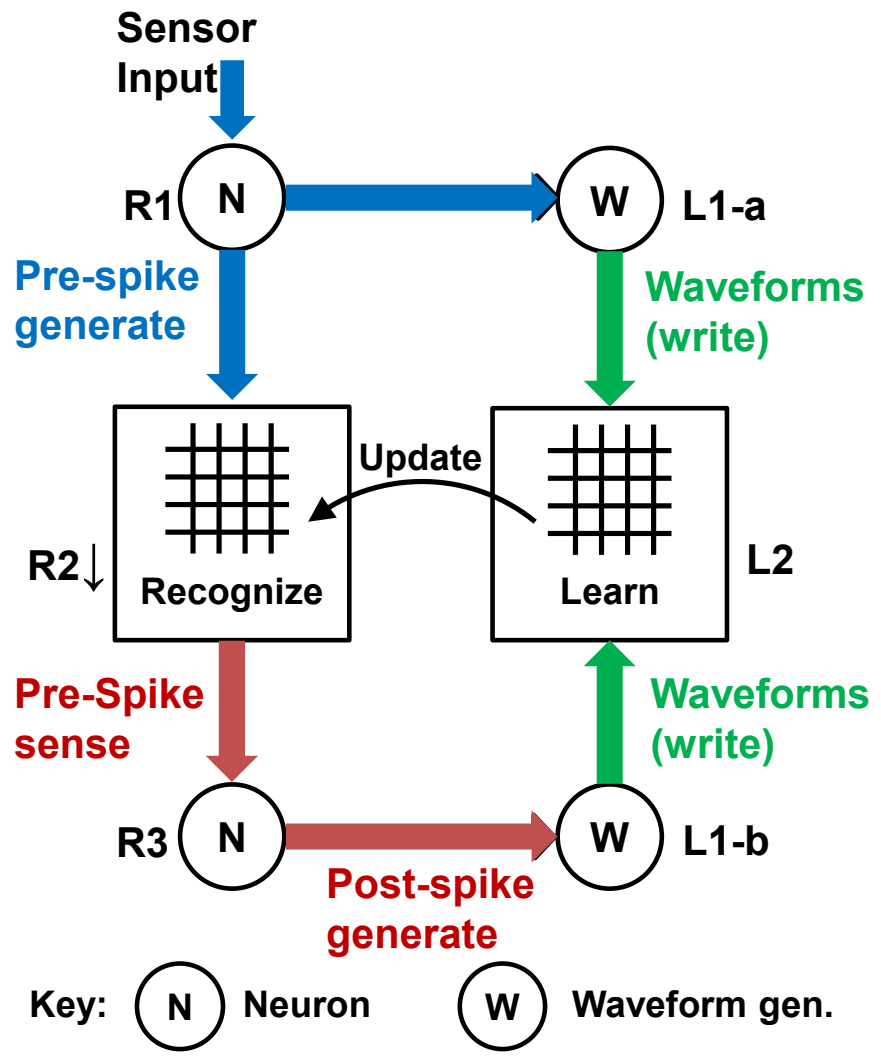

Figure 2: Concept of two array based SNN.

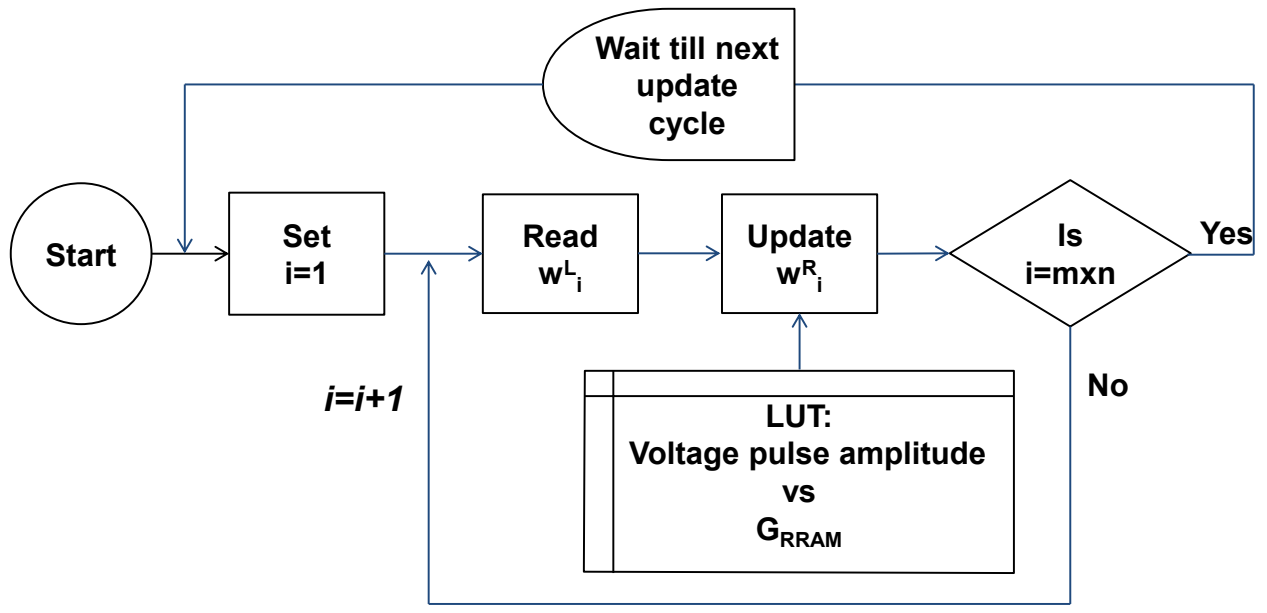

Figure 3: Flowchart for the read-matrix update process, where m and n are matrix dimensions. This process may follow an epoch, after every sample or immediately (Latter-most being ideal for fast learning).

$$\Delta w = \begin{cases} |A_+| \exp\left(-\frac{\Delta t}{\tau_+}\right), & \Delta t \geq 0 \\ -|A_-| \exp\left(-\frac{\Delta t}{\tau_-}\right), & \Delta t < 0 \end{cases} \quad (3)$$

Here, $A_+$, $A_-$, $\tau_+$ and $\tau_-$ are network design parameters.

*e) Recognition and Learning:* Fig. 1(i) shows the electronic-hardware implementation using the same steps as the software. The recognition/reading is done in steps R1 through R3. Step R1 shows an input current proportional to feature-value of sample applied to input neuron. It is converted to voltage spikes (step R2). The scaling of input neurons' spikes can be achieved using an RRAM cross-point array where spike potentials get multiplied with cross-point conductances to give weighted spikes (step R3). These weighted current waveforms then excite and spike the output neurons. The voltage spikes from input neuron (R2) and output neuron (step R3) are fed to wave-form generators for learning (steps L1-a and L1-b). Learning (or, weight change) on an RRAM cross-point array can be achieved using super-position of special waveforms (L2), that are registered in time with the spikes generating them. The superposition of these waveforms gives an above-threshold potential across the RRAM. The time difference between the waveforms determines the net potential that gets applied which, translates to an equivalent conductance change in RRAM (step L3).

### B. Proposed Two Array Scheme

The single array suffers from the recognize (read) vs. learn (program) dilemma. In order to mitigate this challenge, a separate *recognize* and *learn* array is proposed (Fig. 2). In the recognize array, only the read pulses are applied, and in the learn array only the STDP related write-waveforms are applied. Thus, only the learn array synaptic weights are updated in real-time. At regular intervals, the learn array weights are transferred to the recognition array. In a software-equivalent (ideal) scenario, this update should be performed immediately after any weight changes. However, this will slow down learning. Hence more realistic weight updates from learn to recognize array may be performed (i) after a sample is shown or (ii) after every epoch. We will later show the effect of this key difference with SPICE simulations. Fig. 3 shows simple flowchart for updating weights. At a periodic interval, the data input is stopped to "read" the RRAM conductance from learn array to transfer the conductance values to the recognize array. The transfer may be based on look up table of RRAM conductance change vs. applied voltage or using feedback control of multiple small write pulses followed by read until the desired conductance is reached.

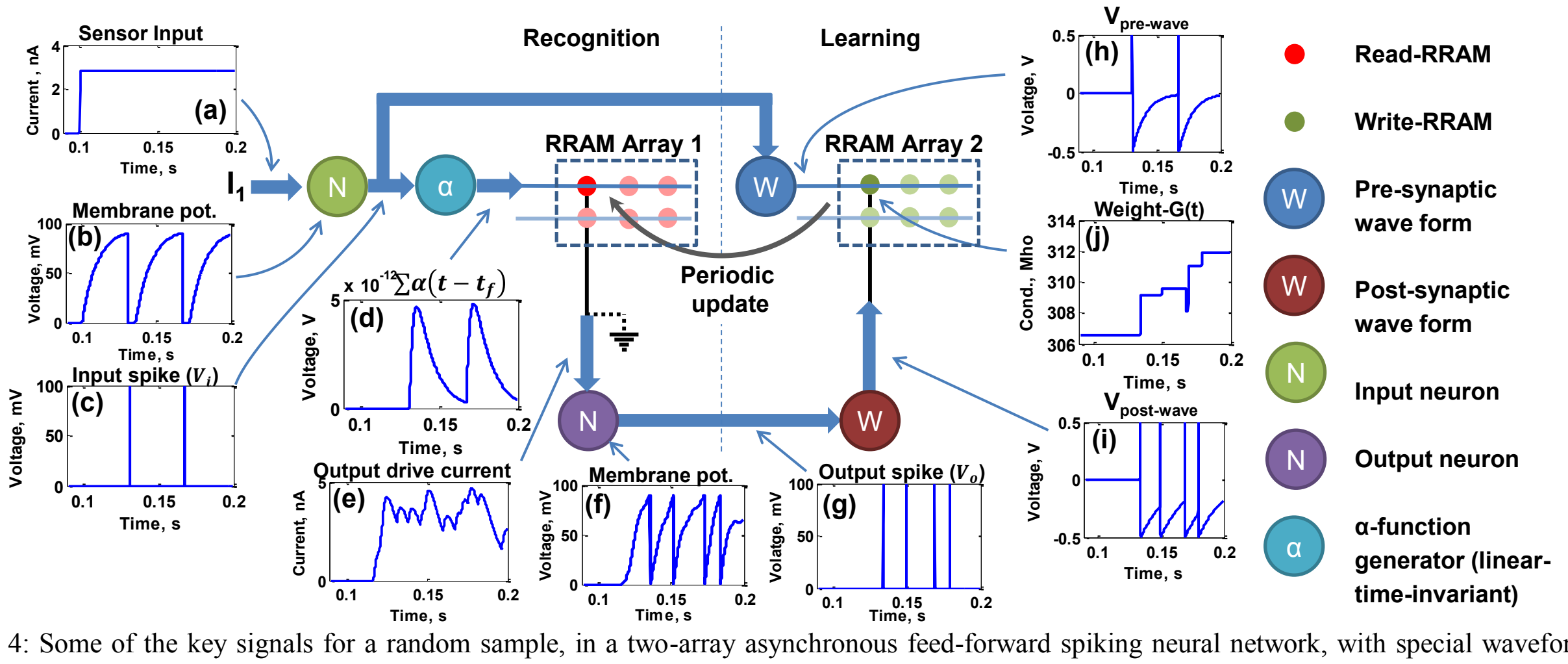

Figure 4: Some of the key signals for a random sample, in a two-array asynchronous feed-forward spiking neural network, with special waveforms for implementing STDP on RRAM; (a) Input-sensing neuron's excitatory current; (b & c) Whenever membrane potential crosses the threshold, input neuron spikes; (d) alpha function, a voltage signal before being weighed and converted into a current signal for the post synaptic neurons; (e) Weighing of alpha-functions by the RRAM array generates excitatory current for the output neuron; (f & g) Whenever membrane potential crosses the threshold, output neuron spikes; (h) waveform generated in response to pre-synaptic spike; (i) waveform generated in response to a post-synaptic spike (Only the most recent spike time determines the waveform signal-value for both cases.); (j) Change in weight of the synapse (or conductance of the RRAM). Anti-causal spikes decrease and causal spikes increase the weight. *Note:* Each of the simulated arrays has a size of 16×3.

### C. Scheme demonstration on SPICE

To validate the proposed scheme, a simple and efficient *Fisher Iris* flower classifier (with 97.33% classification accuracy) given in [18] was designed. This classifier applies four levels of population coding on the four input features of the dataset to produce 4 features×4 levels = 16 input neurons. Thus, the SNN consists of sixteen input-layer neurons connected to three output-layer neurons (corresponding to each class), a 16×3 synaptic weight matrix and exponential STDP of weights. This was implemented in a standard circuit simulator - SPICE. The components developed are briefly described below:

*1) LIF Neuron:* An LIF neuron's core consists of a capacitor (for storing membrane potential), a resistor (for provinding a leakage path) and a reset switch (turns 'ON' after firing). Each time the capacitor's charge crosses a threshold (1) an independent circuit generates a rectangular pulse of constant height and width and (2) capacitor's charge is reset to zero. To incorporate the refractory period, a timer was added that blocks the external current input.

*2) RRAM:* A simplified model was developed for initial analyses. This model consists of a capacitor, whose charge determines the memristor conductivity i.e. synaptic weight. This stored charge can be modified via application of voltage pulses. The software-equivalent ideal RRAM was used for software-equivalence validation of the hardware. Then, realistic RRAM model was used to evaluate effect of realstic RRAM features (e.g. asymmetry in potentiation vs. depression) on the learning performance.

*3) Waveform generators:* This unit is a two-pole linear time-invariant (LTI) system along with a switch. The output potential of this unit is forced to a constant positive voltage with a switch whenever a spike (which is a rectangular voltage pulse) is applied at the input. As soon as the input pulse is removed, the output potential drops to an opposite polarity and subsequently decays (Fig. 1(c)). A new spike would immediately reset and trigger a new waveform.

*4) α-function generator:* This unit is an another LTI system with impulse response given by the *α-function* in Eq. 2. The time constants involved in this equation are much longer than a $V_{spike}$ pulse width, so, the driving spikes can be assumed to be an impulse train. This unit gives sum of time-shifted sum of alpha function in response to incoming spikes (or approximately, train of delta functions).

## III. Results and Discussion

Before carrying out analyses, the components described in SPICE were validated by comparing with software. All software simulations were carried out in MATLAB®.

### A. Component-level validation

*1) LIF Neuron:* As a spike generator is responsible for timing of all the events inside the network, this is the essential part of the becnhmarking procedure. Two kinds of spikes generators were implemented in the SPICE test-bed: (1) with no refractory period, (2) with 5ms refractory period to manage any circuit timing non-idealities. To compare the two neurons (with vs. without refractory period), a uniformly distributed (say, in the range of 0-20nA) random current was applied to emulate a realistic randomly time-varying input current. The output pulses in software vs. hardware were compared. For zero-refractory neuron, 30 spikes were "missing" for 3000 output spikes (1% loss), due to delayed response of SPICE model for same current compared to "instantaneous" software. Larger the observation period, larger is the shift in the last spike, as displacement accumulates over time. In comparison, for 5 ms refractory period, 1 in 294 output spikes (0.3%) went missing. We selected this refractory period for its better accuracy for system level testing. The shift in hardware spikes from their ideal (software) time shows that the average time-period between pulses is 17ms (Fig. 5(a)) while the time off-set is $0 \pm 0.85$ ms (Fig. 5(b)) which corresponds to an error of 5%, which is small.

*2) STDP:* In order to verify execution of the exponential STDP by an ideal RRAM model using waveform superposition, random pulses were applied to the pre- and post-waveform generators, and for each pair of pre- and post-pulse, change in the conductance of the RRAM model was recorded. Fig. 5(c) shows excellent match with the ideal change.

### B. System-level validation

Validation on a system-level is discussed below:

*1) Voltage and Current at various stages:* Fig. 5(d) shows input spikes, synaptic current, output spikes and weight evolution in time. Good registry is observed between software and hardware.

*2) Weight Evolution:* The 16×3 weight matrix evolution at various epochs (initial, $10^{th}$ and $23^{th}$ epoch is shown in Fig. 5(e)) is compared for software vs. hardware. Visually an almost identical pattern can be observed. The correlation between software and hardware is plotted in Fig. 5(f) for same initial condition to show that weight evolution is well-correlated and lie close to 1:1 line In Fig. 5(g), *four* different initial weight distributions are used to observe excellent correlation between software and hardware. Thus we show that ultimately the weights converge to the same final distribution.

*3) Percentage recognition:* Evolution of percentage recognition with training iterations (epochs) has been

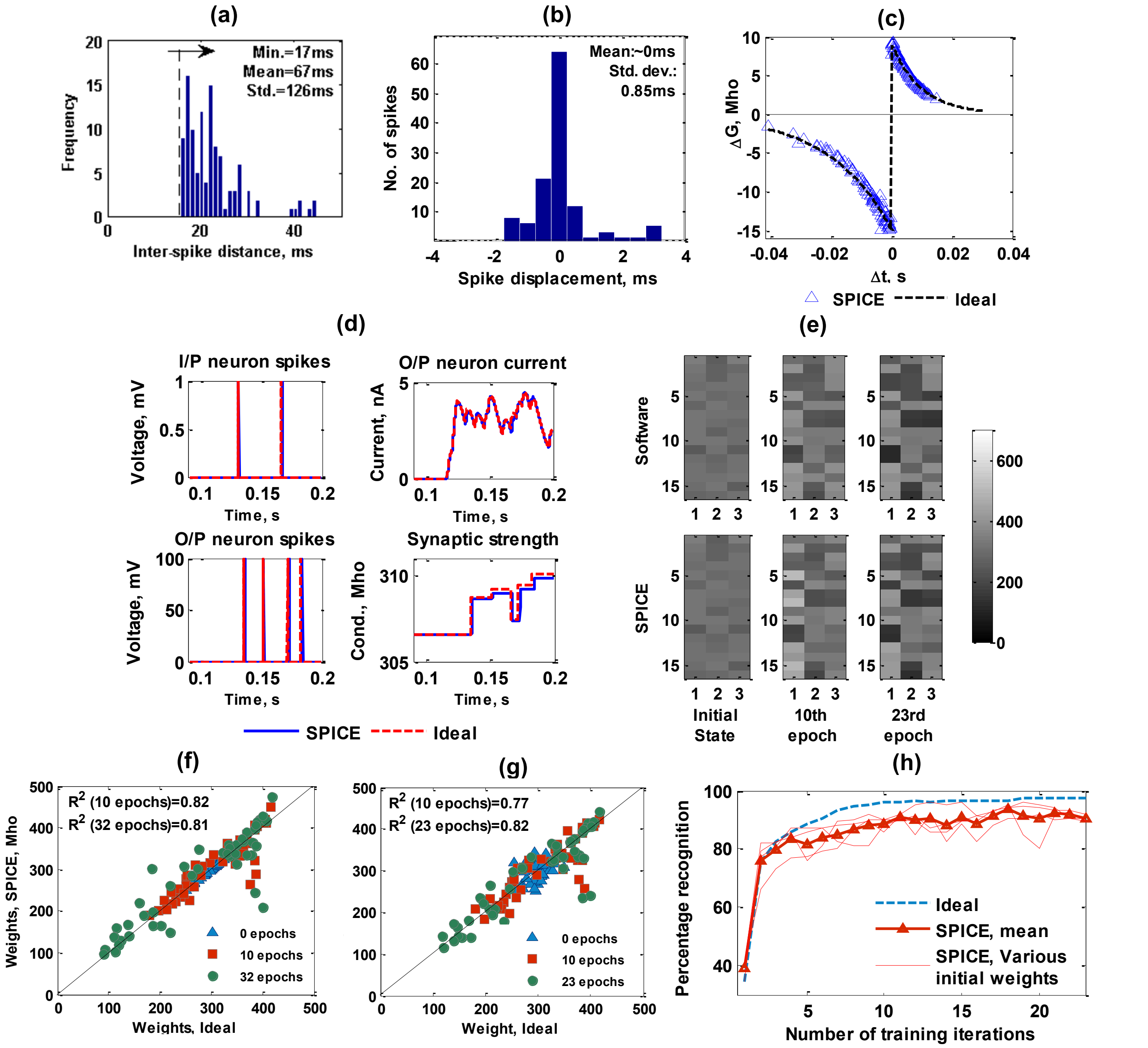

Figure 5: (a) Histogram of inter-spike time from the LIF neuron in SPICE shows minimum 17ms; (b) Comparing of spike times in SPICE with MATLAB as a reference for same excitation current in a 10s window shows very small ($0 \pm 0.85$ ms) shift i.e. 5% of inter-spike distance shows good match between MATLAB and SPICE (c) STDP obtained via waveform superposition on a simplified RRAM model. Comparison of SPICE with software STDP shows that using waveforms, STDP can be accurately achieved; (d) Key signals (input spike, synaptic current, output spike and conductance) evolution in time in SPICE compared with MATLAB is shown. Very little error confirms that SPICE models developed work as desired (e) Color-map of weights during three phases of learning (initial, 10$^{th}$ epoch, 23$^{rd}$ epoch) shows similar weight patterns (f) Scatterplot of weights obtained in SPICE and MATLAB during three phases of learning. Initial weights are set same for MATLAB and SPICE. Value of $R^2$ close to 1 indicates that SPICE based SNN matches software (g) Scatterplot of weights obtained in SPICE and MATLAB during phases of learning. Initial weights are different in this case. With learning iterations, final weight distribution converges to the same distribution indicating insensitivity to initial weight distribution (h) Learning performance in SPICE for four different initial conditions compared with that of MATLAB. Both achieve maximum of 97.3% recognition in all cases.

compared to that of software in Fig. 5(h). Software shows a quick and smooth transition to 97.5% recognition within 10 epochs. The hardware shows an average of 90% recognition but with fluctuations each of the 4 experiments reach the 97.3% recognitions mark. Hence, maximum recognition perfomance is more relevant as each experiment shows such maxima (and minima) during fluctations. So the trainer has the option of stoping the learning at maximum performance. *Hence maximum performance is a more accurate measure of recognition performance* in these cases. Thus, despite a qualitative difference (somewhat more noisy), the recognition performance is excellent.

## C. *Effect of Realistic Weight Update schemes*

As discussed, the weights of the learn array are updated naturally real-time. However, the learn-array weights need to be transferred/ updated to the recognition array periodically. In the soft-ware equivalent bench-marking, an unphysical instantaneous update was used to demonstrate excellent performance. Here we compare this with two more realistic updating scheme for recognition array (1) post-sample update (period for update is same as the training sample application period) (3) inter-epoch update (weights are updated after all the training samples are input to complete at epoch). The percentage recognition with training iterations (epochs) is compared for each of these schemes in Fig. 6. No significant difference between the three update schemes is observed and hence excellent recognition performance is observed.

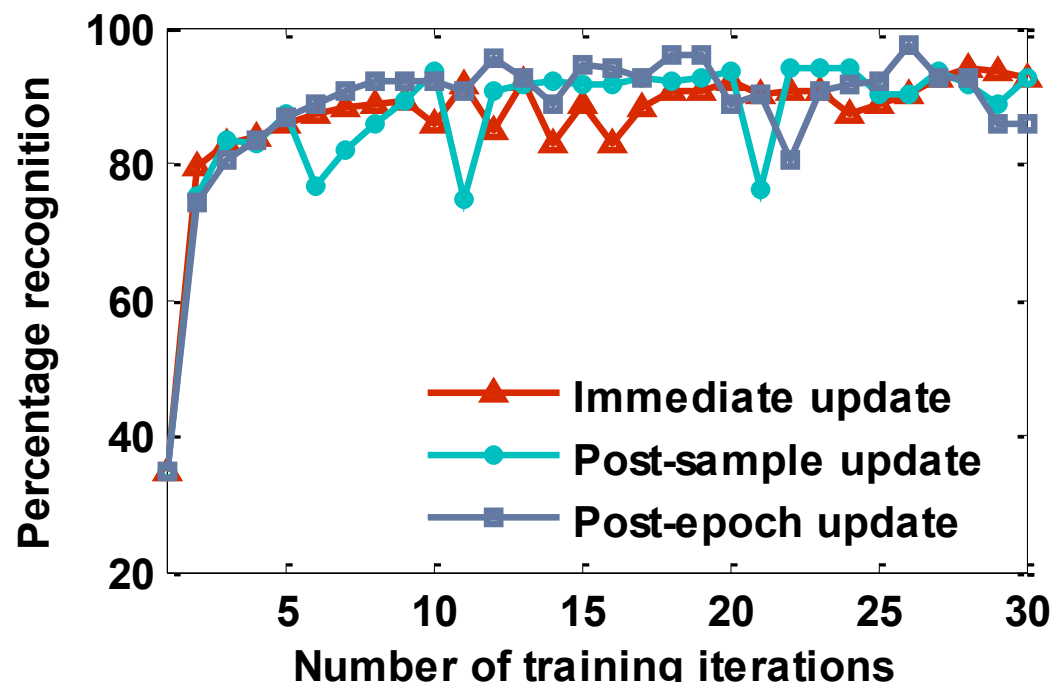

Figure 6: Learning performance for various update-schemes (same initial weights used.)

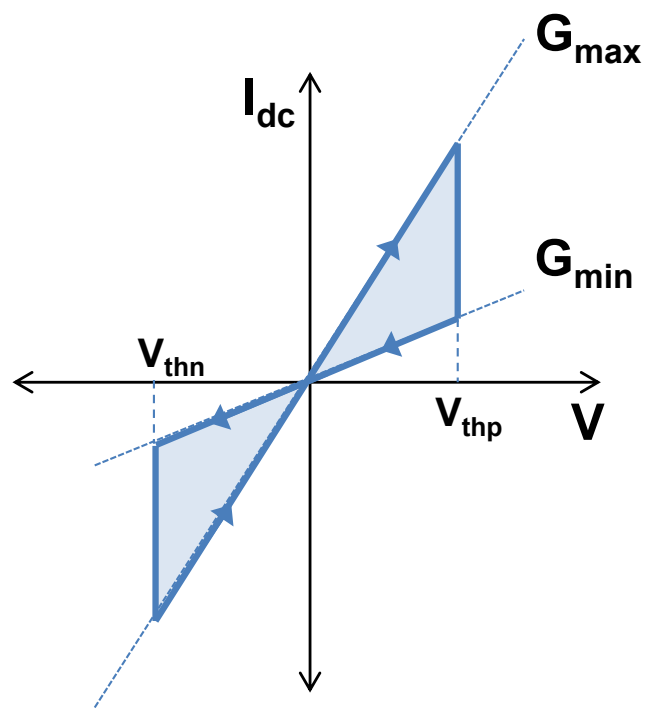

Figure 7: Linear DC-IV of an ideal-RRAM. It behaves as an ideal resistance, with no change in threshold, unlike a real RRAM.

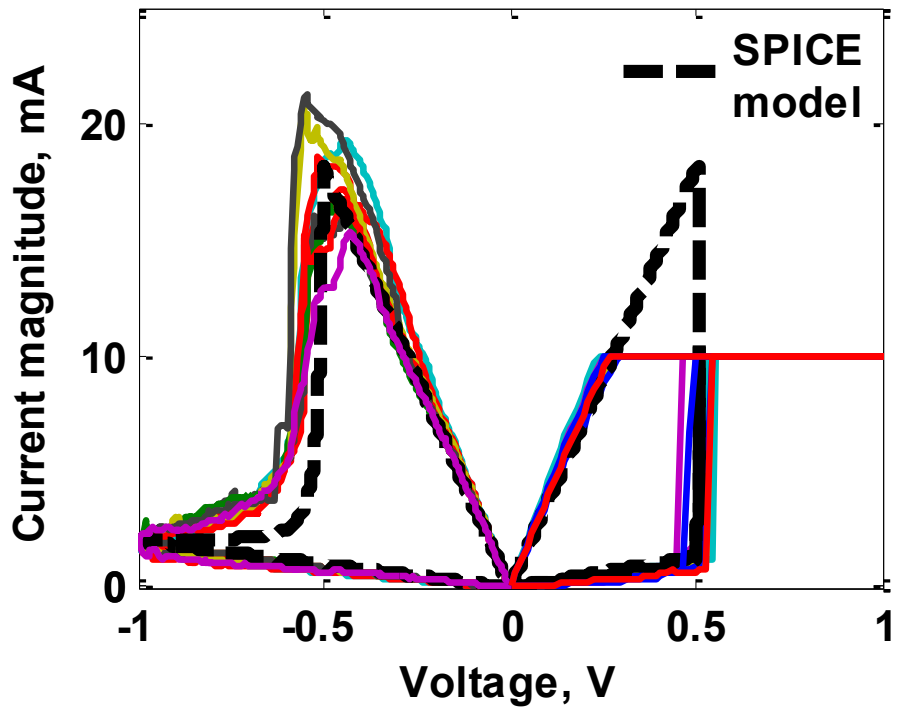

Figure 8: *Colored curves:* Experimental DC-IV of an RRAM. There is negligible change in threshold on the positive side, and large change on the negative side (esp. where resistances are high); *Dashed curve:* SPICE model's DC-IV. For this RRAM, $G_{max}/G_{min} \approx 25$.

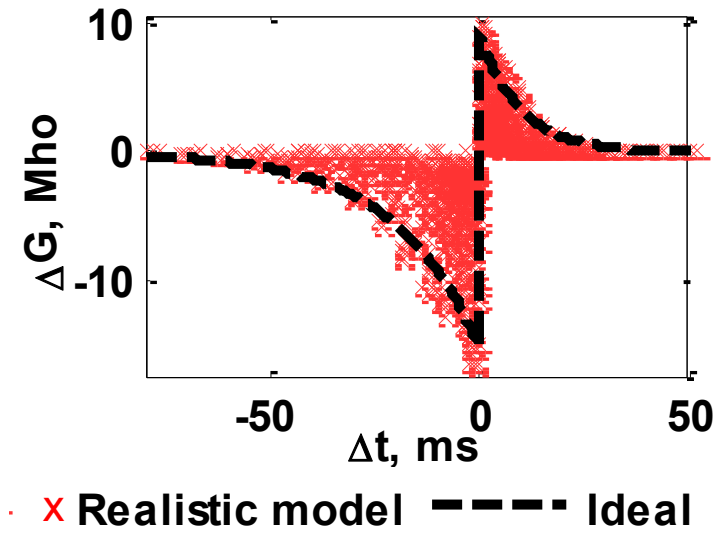

Figure 9: STDP, as shown by a realistic RRAM model. Notice that it goes to zero, due to the incorporation of saturation effect into the model (i.e., change in conductance slowly goes to zero as conductance attains a certain high or low value). Similar trend was originally observed in rat's Hippocampal neurons (See Fig. 10).

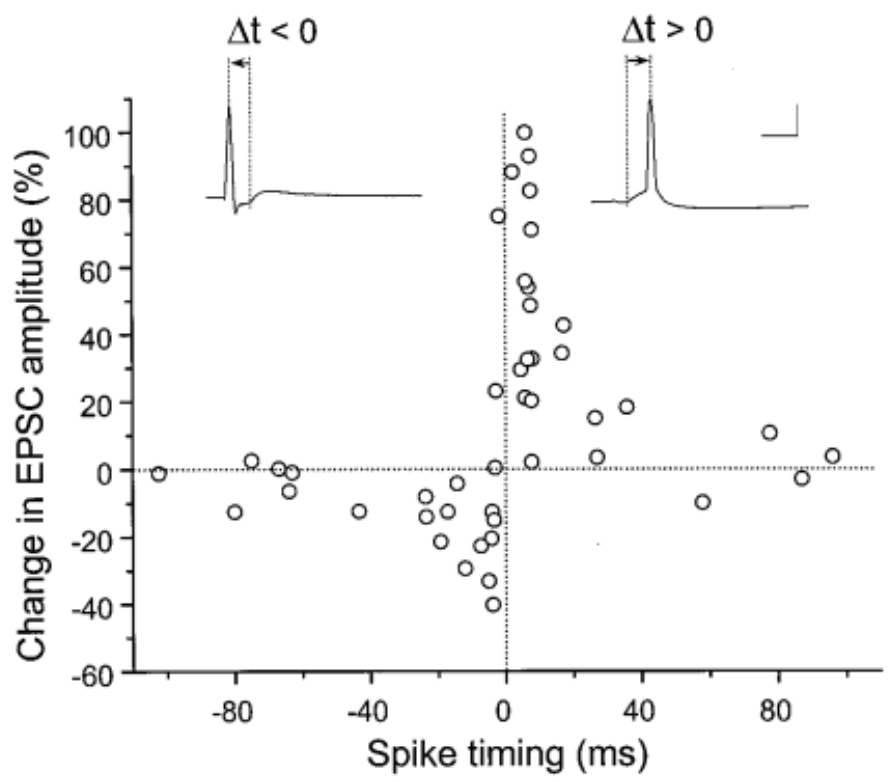

Figure 10: STDP, as experimentally discovered by Bi et al. [20] in a rat's Hippocampal neurons. Significant LTP was observed in synapses with low initial strength. LTD did not have any obvious relation with initial strength.

### D. *Effects of realistic RRAM*

The ideal RRAM IV characteristic shows "abrupt" and symmetric set (high to low resistance change) & reset (vice versa) as shown on Fig. 7. In comparison, experimental switching characteristics of $HfO_2$ RRAM [19] show more asymmetric and voltage-dependent set/reset. To study the effect of a realistic RRAM, a circuit model with switching threshold varying with its conductance was developed to match experimental IV characteristics. Fig. 8 shows good match between experiment and circuit model and Fig. 9 shows STDP observed with the model by applying randomly timed spikes. This realistic RRAM model was then used for real-time learning. The qualitative learning performance evolution with training is shown in Fig. 10. Though a qualitative similarity is observed including fluctuation, the realistic RRAM model achieved lower maximum (85%) and average (75%) recognition performance. However, due to fluctuations, a maximum fluctuation is a better measure as justified earlier.

This also implies that RRAM engineering is required for software-equivalent hardware. This set-up is an excellent test for experimentally-validated RRAM models.

## IV. Discussion and Benchmarking

A summary of results is shown in Table I. First, completely software-equivalent hardware is demonstrated with excellent performance – even through there were small microscopic and macroscopic deviations from the ideal (software). Second, realistic weight transfer/update scheme from learn to recognize array were explored to show that the hardware is robust to infrequent updates like post-epoch update scheme. Finally, a realistic RRAM model was developed and matched to experiment. Though the performance was still quite good, slight degradation in performance was observed compared to ideal RRAM model. This implies that RRAM engineering is critical for RRAM (or memristor) based SNN hardware.

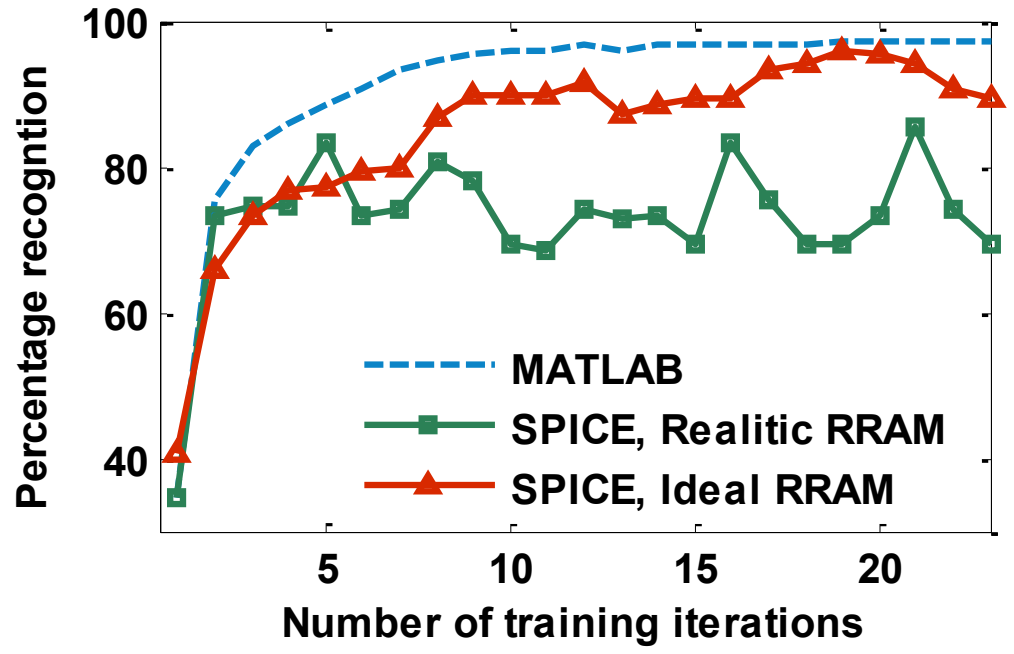

Figure 10: Learning performance for realistic RRAM model. The maximum recognition percentage goes up to 85%.

TABLE II. SUMMARY OF LEARNING PERFORMANCES

| | | ***Percentage recognition*** ***Maximum** (mean)* |
|---|---|---|
| MATLAB | | **97.3%** (97.3%) |
| SPICE | Immediate update | **97.3%** (90%) |
| | Post-sample update | **97.3%** (90%) |
| | Post-epoch update | **97.3%** (88%) |
| | Realistic RRAM model | **85%** (75%) |

Table 2 shows the specific improvements demonstrated in this work. The work demonstrates an asynchronous system compared to digital and clocked systems [11], [13]. Carlos Zamarreño-Ramos et al. [15] also propose an asynchronous scheme a *long* wave-form is used instead of a *sharp* spike to enable read and write (for STDP). Further, the post neuron integrates spikes, except when it is generating waveform of its own, during which it *stops integrating* current. Thus it addresses the *simultaneous read-write dilemma,* but ignores the input during specific period after the post neuron fires. This post spike related temporary inhibition of integration is argued to be similar to the refractory behaviour observed in biological neurons. However, the length of the refractory period is a design parameter in algorithms (which may even be *zero*), independent of the waveform time-scale in general, which is determined by STDP time-scale. In comparison, in this work, sharp pulses are used for recognition in the recognize-array, while the learning using special waveforms is performed on a separate learn-array in the two-array scheme. An excellent software-equivalence of hardware is shown to demonstrate that by this scheme any SNN can be directly converted to hardware without the need to adapt the system to a digital signalling scheme or adapt circuit components e.g. neuron etc. to non-biomimetic waveforms for recognition.

## V. CONCLUSIONS

In conclusion, a SPICE based software-equivalent SNN hardware using two independent RRAM arrays was proposed and demonstrated through a Fisher Iris dataset classifier. The maximum percentage recognition obtained was found to be robust and comparable to software performance – despite some differences in hardware scheme (e.g. periodic weight transfer and some small time differences of spikes compared to software). Secondly, a realistic RRAM circuit model matched to experimental IV characteristics was compared with ideal RRAM to show a slight degradation in performance. This highlights the need to engineer RRAM behavior specifically for SNN applications.

TABLE I. SCHEMES FOR LEARNING ON SPIKING NEURAL NETOWORK HARDWARE

| | ***Key features*** | | | | | |
|---|---|---|---|---|---|---|
| | ***Learning algorithm/rule*** | ***Change in weight*** | ***Synapse*** | ***Simultaneous Read- write*** | ***Weight update cycle*** | ***Learning vs. recognition*** |
| Seo et al. [11] | General STDP | Probabilistic set/reset | 8T-SRAM | Yes | Next clock cycle after spike | Multiple Clock phases are used for independent learning and recognition phases |
| Y. Kim et al. [13] | General STDP | Continuous using Pulse Width Modulation | Idealized memristor model | No | Next clock cycle after spike | |
| Carlos Zamarreño-Ramos et al. [15] | General STDP | Continuous using waveform superposition | Idealized memristor model | Yes | Asynchronous update | Asynchronous waveforms used for both learning and recognition; Refractory period is long & fixed. |
| This work | General STDP | Continuous using waveform superposition | Realistic memristor / RRAM model | Yes | Asynchronous with Post-sample, post-epoch weight transfers | Pulse (biomimetic) is used for recognition and waveform for learning |